\documentclass{Interspeech2024}

\interspeechcameraready 
\usepackage{multicol}
\usepackage{multirow}
\usepackage{tablefootnote}
\usepackage{amsmath}
\usepackage{subcaption}

\title{Multi-aspect Depression Severity Assessment via Inductive Dialogue System}

\name[affiliation={1}]{Chaebin}{Lee}
\name[affiliation={1}]{Seungyeon}{Seo}
\name[affiliation={1}]{Heejin}{Do}
\name[affiliation={1, 2}]{Gary Geunbae}{Lee}

\address{
  $^1$Graduate School of Artificial Intelligence, POSTECH, South Korea\\
  $^2$Department of Computer Science and Engineering, POSTECH, South Korea}
\email{\{leecbin0911, ssy319, heejindo, gblee\}@postech.ac.kr}


\begin{document}
\maketitle

\begin{abstract} 
With the advancement of chatbots and the growing demand for automatic depression detection, identifying depression in patient conversations has gained more attention. 
However, prior methods often assess depression in a binary way or only a single score without diverse feedback and lack focus on enhancing dialogue responses. In this paper, we present a novel task of multi-aspect depression severity assessment via an inductive dialogue system (MaDSA), evaluating a patient's depression level on multiple criteria by incorporating an assessment-aided response generation. Further, we propose a foundational system for MaDSA, which induces psychological dialogue responses with an auxiliary emotion classification task within a hierarchical severity assessment structure. We synthesize the conversational dataset annotated with eight aspects of depression severity alongside emotion labels, proven robust via human evaluations. Experimental results show potential for our preliminary work on MaDSA.
\end{abstract}

\section{Introduction}
As stated by the World Health Organization (WHO), depression is a prevalent mental health disorder in contemporary society, affecting around 5\% of the global adult population.
However, diagnosing depression is challenging due to its intricate nature of being influenced by various factors. To automatically diagnose depression, many researchers have delved into automatic depression detection \cite{microblog, twitter1, twitter2}. 
Specifically, with the rising prominence of chatbots and the prevalence of daily conversations, various studies identify depression using conversation data \cite{daicwoz, bindepression1, textbasedet2}. 

Existing studies typically handle the task as a binary classification \cite{microblog, twitter1, bindepression1} (e.g., whether someone has depression or not) or predict a single overall score \cite{regdepression1, regdepression2}. Moreover, these studies often do not focus on interactively leveraging dialogue response generation, only assessing the depression in the given text.
A simple binary approach and holistic severity prediction could not provide detailed quantitative evidence for the diagnosis of multifaceted depression, resulting in a lacked explainability and reliability, which are essential in medical AI. 
Furthermore, the absence of eliciting intended response leads to missing potential clues or signals that could aid in diagnosis, thereby deviating from the real-world situations where depression is diagnosed within a natural context. 

In this paper, we propose a novel task, multi-aspect depression severity assessment via inductive dialogue system (MaDSA), which measures a patient's depression severity at multiple criteria such as \textit{Interest}, \textit{Fatigue}, and \textit{Self-esteem}, within natural conversational situations. In particular, we introduce a fundamental system that generates contextual responses and induces users to answer to the psychological response, Patient Health Questionnaire–8 (PHQ-8) \cite{phq8}, which has eight distinct aspects for depression severity assessment. 
For the assessment, we employ a hierarchical structure that first captures the turn-level and then dialogue-level representations, inspired by the automated writing evaluation tasks \cite{ridelytrait, heejin}. In addition to determining the presence of depression, our multifaceted assessment could provide tailored feedback by pinpointing specific areas contributing to its severity.

Emotional interaction is another crucial element in the context of depression conversation agents, unlike general dialogue systems \cite{emotionalGen1, emotionalGen2}. Therefore, we employ an auxiliary task to classify emotions as positive or negative based on the user's input. In particular, when the model-predicted emotion is negative, we gauge contextual similarity between the prior dialogue vector and the candidate vectors, including model-generated and PHQ-8 question vectors, to induce user response for fine-grained assessment. This procedure of capturing emotional nuances in generating psychological responses facilitates comprehensive assessment. 

Given the absence of a multi-aspect-labeled dataset, we create the synthetic dialogue dataset labeled by the PHQ-8 aspects (scale 0--3) for the MaDSA task. By conducting human evaluation with three certified English-proficient psychology-domain experts, we confirmed a significant correlation between the labels in our data and those tagged by the experts. Experimental results and ablation studies to investigate the impact of model components reveal that jointly generating inductive questions with predicting emotions enhances the assessment quality. Additional examination to perform the depression identification task with the DAIC-WOZ \cite{daicwoz} further proves the generalizability of our method, showing a 2.48\% improvement in the accuracy. This result underscores the value of measuring severity levels across multifaceted aspects, thereby aiding in the diagnosis of depression.

\section{Related Work}

With the rising demand for diagnosing depression, many researchers have explored the realm of automated depression detection. For instance, some studies \cite{microblog, twitter1} have focused on leveraging social media posts as a basis for automatic depression detection. As the importance of popularity grows and daily conversations become more common, using conversation data for depression classification is also becoming essential. For this purpose, DAIC-WOZ \cite{daicwoz}, a multi-modal conversation dataset, was introduced in AVEC 2016 \cite{avec}. Although it provides rich dialogues annotated with indicators of depression and its severity, holistic labels without multi-aspect-wise scores overlook explanatory factors for depression severity. Accordingly, prior research efforts focused on classifying depression in a binary manner \cite{microblog, twitter1, bindepression1} or predicting a single overall severity \cite{regdepression1, regdepression2}, lacking a nuanced exploration of the detailed depression components. Distinguished from them, we propose a new framework that surpasses conventional, holistic diagnosis by enabling severity assessment across multiple dimensions.

Emotion classification has been utilized for generating contextually appropriate responses \cite{emotionalGen2, mutlitaskGen} and contributes to improving the depression severity assessment performance \cite{emotionDetection1, multitaskdetection}. As emotions provide valuable insights into one's mental state, they can be an explicit clue for depression detection \cite{zhang2021multi, prabhu2022harnessing}. Moreover, considering that depression is associated with impaired emotion regularization, capturing and understanding emotion is a core element of accurate depression diagnosis \cite{theorical1, theorical2}.
Motivated by prior works and theoretical basis, we leverage emotion classification as an auxiliary task in which the model could better capture the emotional aspects. 

\section{MaDSA}

\begin{figure}[t]
    \centering
    \includegraphics[width=\columnwidth]{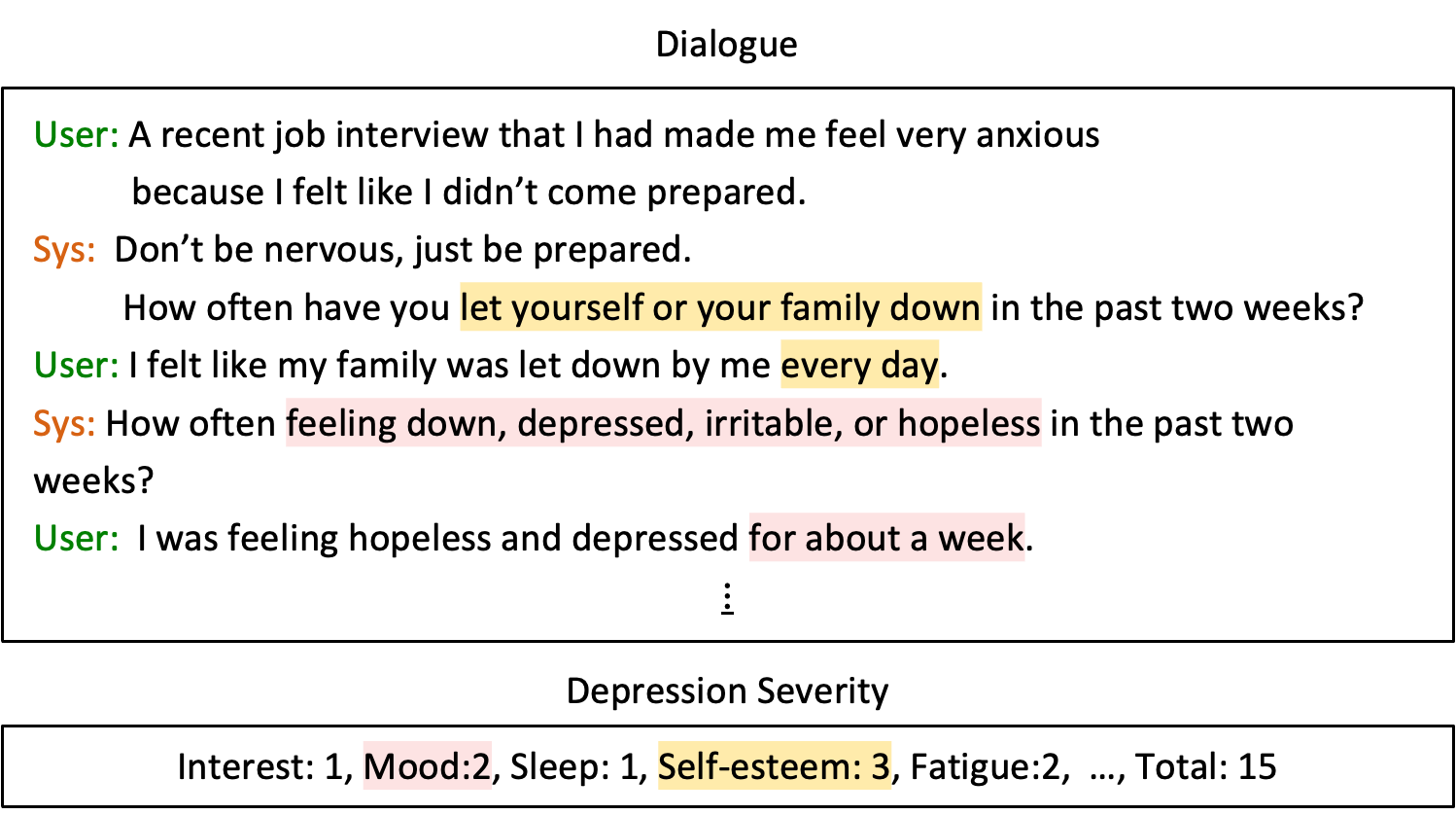}
    \caption{Example of the synthetic data. The yellow represents self-esteem items and the pink represents mood items.}
    \label{fig:data}    
\end{figure}

\subsection{Dataset Synthesis}\label{sec:data}

Given the absence of conversational data that includes both emotion labels and annotations with PHQ-8 scores, we create a synthetic dataset based on the DailyDialog dataset \cite{dailydialog} and EmpatheticDialogues (ED) \cite{empathetic}. They are multi-turn open-domain English dialogue data with emotion labels. Our primary goal is to quantify multi-faceted depression levels within the conversational data. Note that the response for the PHQ-8 questionnaire is an integer ranging from 0 to 3; thus, converting these numerical scores into natural and text-based responses is required to integrate the user's symptoms and their score into the dialogue.

For this, we train T5 model for user-response-generation (URG) that generates diversely paraphrased user responses for each score label. We pre-define multiple anticipated user answers based on each of the PHQ-8 questions regarding eight aspects of \textit{Interest}, \textit{Mood}, \textit{Sleep}, \textit{Appetite}, \textit{Fatigue}, \textit{Self-esteem}, \textit{Concentration}, and \textit{Moving}.
Then, the URG model is trained to produce the user response given the PHQ-8 question and its corresponding score input. Considering that modifying PHQ-8 questions could potentially change their psychiatric properties, we primarily focus on diversifying user responses.

The URG model is trained with the negative log-likelihood:
\begin{small}\begin{equation}\label{eq:para_loss}
    L = -\frac{1}{NT}\sum_{n=1}^{N} \sum_{t=1}^{T} \log P(u^n_{t} |  q^n_{t} ; s^n_{t} )
\end{equation}\end{small}
where the model is fed with input, including a PHQ-8 question $q_t^n$, and its score $s_t^n$ within the $t$-th turn of the $n$-th dialogue; this generates a user response $u_t^n$.
Then, using the data sample labeled with \textit{negative} emotion in the DailyDialog and ED dataset, we calculate the similarity between the conversation history vector and system responses candidate vectors that include both PHQ-8 questions and the response of the original dataset for each turn. Based on this computation, the system's response is determined as the most similar candidate to the historical context. When the selected response is one of the PHQ-8 questions, the trained URG model generates the subsequent user response corresponding to the question and its aspect-specific score.
In the case of \textit{negative} emotion label, the final system response can be formulated as follows: 

\begin{equation}
\label{eq:argmax}
\begin{split}
    r_\textnormal{syn} = \textnormal{argmax}_{r \in r_\textnormal{cand}}(\mathrm{sim}(r,c)) \\
    \textnormal{where}~r_\textnormal{cand} = r_\textnormal{phq} \cup r_\textnormal{original}
\end{split}
\end{equation}
where $r_\textnormal{syn}$, $r_\textnormal{cand}$, and $c$ denote the synthesized system response, response candidates, and history context vector, respectively. Here, we use the cosine similarity for the $\mathrm{sim}(\cdot)$ function. Consequently, we made 95,287 training and 13,078 development samples. Figure~\ref{fig:data} shows an example of the generated data.

\begin{figure}[t]    
    \centering
    \includegraphics[width=\columnwidth]{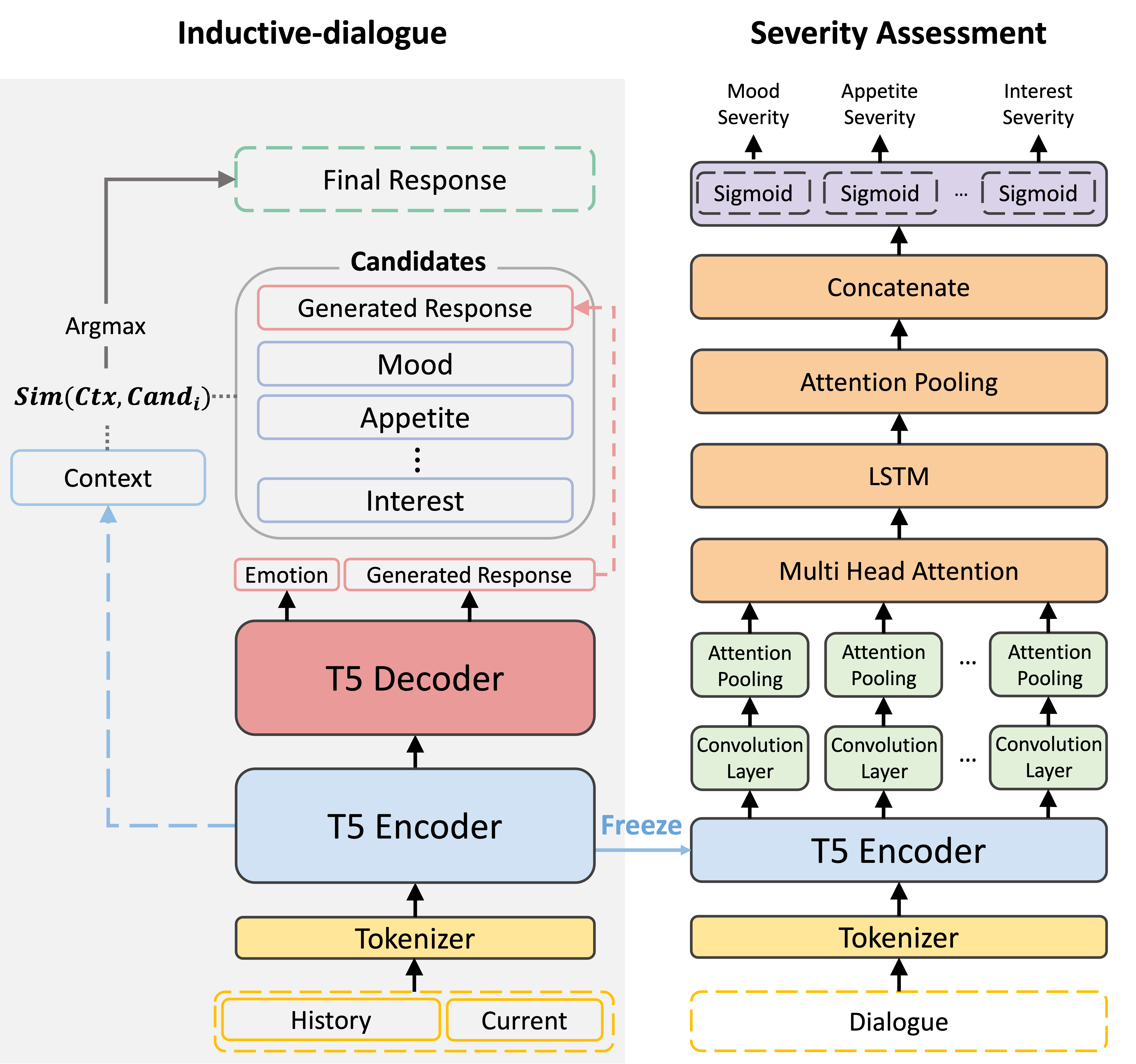}
    \caption{The left is an inductive dialogue system, and the right is a severity assessment system trained on a frozen encoder of the dialogue system.}
    \label{fig:arch}    
\end{figure}

\begin{table*}[t]
\centering
\caption{\label{tab:qwk}
Results of depression severity assessment for each aspect. Each value denotes the QWK score. QI and EC denote \textit{Question Induction} and \textit{Emotion Classification}, respectively. The \textbf{bold} text identifies the highest value of the aspect.}
\scalebox{
0.9}{
\begin{tabular}{l|cccccccc|c}
\hline
& \multicolumn{8}{c|}{\textbf{Aspects}} & \\
\hline
\textbf{Model} & Interest & Mood & Sleep & Appetite & Fatigue & Self-esteem & Concentration & Moving  & Average \\
\hline
T5-baseline & 0.858 & 0.883 & 0.857 & 0.846 & 0.908 & 0.843 & 0.852 & 0.786 & 0.866 \\
\hline
+QI & 0.849 & 0.894 & 0.839 & \textbf{0.850} & 0.906 & {0.860} & 0.850 & 0.794 & 0.867 \\
+EC & 0.861 & 0.891 & 0.850 & 0.839 & 0.916 & \textbf{0.863} & 0.853 & \textbf{0.820} & 0.873 \\
 MaDSA (+QI +EC) &\textbf{0.869} & \textbf{0.905} & \textbf{0.860} & 0.846 & \textbf{0.920} & 0.856 &\textbf{0.856} & 0.807 & \textbf{0.877}\\
\hline
 \end{tabular}}
\end{table*}

\subsection{Inductive Dialogue System}
\label{sec:ids}
For the baseline dialogue system, we employ the pre-trained T5 \cite{T5}\footnote{\url{https://huggingface.co/t5-small}}, which auto-regressively generates the response, to ensure that the produced sentences encompass emotion-related information cohesively. To generate contextually appropriate and assessment-aided responses, the system conducts emotion classification for each dialog turn. Based on this classification result, we follow a methodology similar to the data synthesis. In cases where positive emotions are detected, the system proceeds with a standard open-domain conversation flow. Conversely, if the negative emotions are identified, the system generates the final response by calculating the cosine similarity between history context and candidate vectors, consisting of model-generated and PHQ-8 question vectors, and the conversation-history vector (the similar process in the Equation \ref{eq:argmax}). 
This procedure allows the model to generate PHQ-8-related questions within natural contexts, enhancing the quality of depression assessment without harming the flow of the overall conversation. The following negative log-likelihood loss is used to train the model for emotion and response generation: 
\begin{small}\begin{equation}
    L_{\textnormal{\scriptsize{gen}}} = - \frac{1}{NT}\sum_{n=1}^N \sum_{t=1}^T \log p(e_{t}^n; s_{t}^n| h_{t}^n )
\end{equation}\end{small}
where $e_t^n, s_t^n, u_t^n$, and $h_t^n = \{ u_1^n, s_1^n, \dots, u_t^n\} $ denotes emotion, system response, user response, and conversation history at $t$-th turn of $n$-th dialogue, respectively.
On the left side of Figure~\ref{fig:arch}, the entire process of the response generator with inducing questions is illustrated.

\subsection{Depression Severity Assessment}
To assess depression severity in a comprehensive and multi-aspect manner, we have repurposed ProTACT \cite{heejin}, which was initially devised for the automated writing evaluation (AWE) tasks. 
Given the importance of contextual comprehension in our task, we employ the encoder from the previously trained dialogue generator as our chosen embedding method instead of the part-of-speech embedding that is widely used in the AWE.
The input dialogue is first embedded at the word level using the encoder of the inductive dialogue system (Sec~\ref{sec:ids}) with frozen weights. Then, it is passed through a 1D convolutional layer, which is followed by an attention-pooling layer to access the turn-level information. The process can be defined as follows:
\begin{eqnarray}
    \mathbf{c}_i &=& f(\mathbf{W}_z \cdot [x_i : x_{i+ h_w - 1}] + b_z) \\
    \mathbf{t} &=& \sum \alpha_i \mathbf{c}_i \label{eq:att3}
\end{eqnarray}
where $\mathbf{W}_z$ represents the trainable weights matrix of convolution; $b_z$ is a bias vector; and $h_w$ is the size of the convolutional window. The turn-level $\mathbf{t}$ is achieved by taking a weighted sum of the representation  $\mathbf{c}$ from the convolution layer and attention vector $\alpha_i$. The last level, dialogue-level representation, can be obtained as follows:
\begin{small}\begin{eqnarray}
    H_{i}^{j} &=& \textnormal{Attention} (T W_i^{j1}, T W_i^{j2}, T W_i^{j3}) \\
    \textnormal{m}^i &=& \textnormal{concat} (H_i^1, H_i^2, \dots,  H_i^h) W^j \\
    h_t^i &=& \textnormal{LSTM} (m_{t-1}^i, m_t^i)
\end{eqnarray}\end{small}
When grading the $i$-th item, the output $T$ from the previous layer, converted into a matrix of turn-level representations, is used as the query, key, and value for the scaled-dot product attention. 
Specifically, $H_i^j$ represents the $j$-th head of the $i$-th item from the multi-head attention, and $h_t^i$ denotes the hidden representation for the $i$-th item at time-step $t$ from the recurrent layer LSTM \cite{lstm}. Finally, the score is predicted using the sigmoid function after the attention pooling. 
The model is trained with the mean-squared error.
As the sigmoid function predicts the score, the scaled output score prediction is reconverted to its original scale in the inference phase. The right side of Figure~\ref{fig:arch} illustrates the overall process of depression severity assessment.

\section{Experiments}
\subsection{Settings}
We utilize a pre-trained T5 model as the tokenizer and baseline for the inductive dialogue system. We set the input maximum length as 512, the maximum turn length considered from the history as 7, the epoch as 10, and the batch size as 16. We employ the AdamW optimizer \cite{admaw} with a learning rate of 1e-5. For the assessment system, during 20 epochs, we set LSTM units as 100, CNN filters as 100, and CNN kernel size as 5, with a dropout rate of 0.3.

\subsection{Evaluation Metrics}
To evaluate the quality of the generated response, we specifically use two variants of the BLEU metric \cite{bleu}, BLEU-1 and BLEU-2. For depression severity assessment, we employ the quadratic weighted kappa (QWK) score, Widely used in evaluation indicators in the AWE field \cite{ridelytrait, heejin, trait1}. The QWK measures the extent of agreement between the predicted and human-rated scores, and is defined as follows: 
\begin{small}\begin{eqnarray}
    W_{y,\hat{y}} = \frac{(y-\hat{y})^2}{(R-1)^2} \\
    K = 1 - \frac{\sum W_{y, \hat{y}} O_{y, \hat{y}}}{\sum W_{y, \hat{y}} E_{y, \hat{y}}}
\end{eqnarray}\end{small}
where $y$, $\hat{y}$, and $R$ are the target score, predicted score, and number of possible scores, respectively. An $R\times R$ matrix $O$ has a component  $O_{y,\hat{y}}$ that denotes the frequency score of $y$ and $\hat{y}$. The expected score matrix $E$ is calculated by the outer product between histogram vectors of $y$ and $\hat{y}$.

\section{Results}
\subsection{Main Results}
Table~\ref{tab:qwk} shows the QWK results of the depression severity assessment for each aspect on the synthesized data. We use T5 as the baseline because it is the representative pre-trained language model comprising both an encoder and decoder. Our approach, combining both emotion classification (EC) and question induction (QI), outperforms the baseline. Notably, the performance enhancement is more substantial when \textit{QI} and \textit{EC} are employed simultaneously than when applied independently. This outcome highlights the synergistic effect of auxiliary emotion classification and question induction, emphasizing the significant role of inductive dialogue in improving the accuracy of severity assessment for multiple aspects.

We further explore the ablation results by individually examining the impacts of adding \textit{QI} and \textit{EC} (Table~\ref{tab:qwk}). Notably, solely applying emotion classification significantly improves emotion-related aspects like \textit{Interest}, \textit{Mood}, \textit{Fatigue}, and \textit{Self-esteem}. Predicting the emotional class within the assessment system might lead to obtaining emotion-related representation. 
The results emphasize that our auxiliary task could implicitly assist the depression severity scoring.

\begin{table}[t]
\centering
\caption{Results of dialogue generation.}
\scalebox{0.8}{
\begin{tabular}{l|c|c}
\hline
     & BLEU-1 & BLEU-2 \\ 
     \hline
T5-baseline &  1.103  & 0.960    \\ \hline
+QI     & 0.985  & 0.825       \\ \hline
+QI +EC (MaDSA)  & {1.073}  & 0.932  \\ \hline
\end{tabular}}

\label{tab:dialGen}
\end{table}

\subsection{Effects on Response Generation} 
To ensure a multi-aspect assessment for depression severity, our dialogue system generates the PHQ-8 questionnaire; however, it might blur the context and reduce the diversity of the dialogue system. Consequently, when we include the questionnaire input, the BLEU score decreases (Table~\ref{tab:dialGen}). Nevertheless, when combining emotion classification and question induction, the BLEU scores are improved and comparable to the baseline. This result suggests that predicting emotion during inductive response generation allows the model to preserve contextual coherence.

\begin{figure}[]
	\centering
        \subfloat[]{
        \vspace{-\abovecaptionskip}
	\begin{minipage}{.74\columnwidth}
		\centering
            \includegraphics[width=\columnwidth]{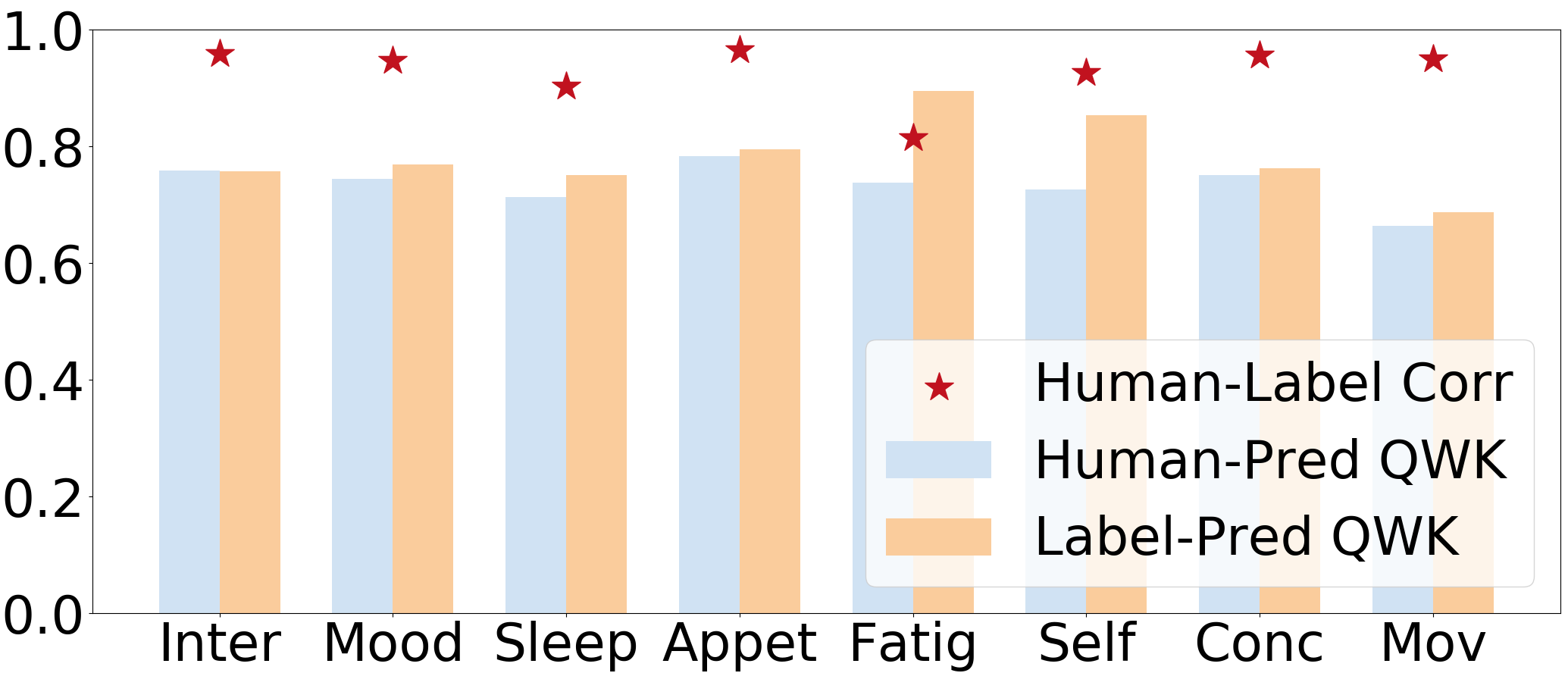}
		\label{subfig:qwk}
	\end{minipage}}
        \hfill
        \subfloat[]{
        \vspace{-\abovecaptionskip}
	\begin{minipage}{.24\columnwidth}
		\centering
            \includegraphics[width=\columnwidth]{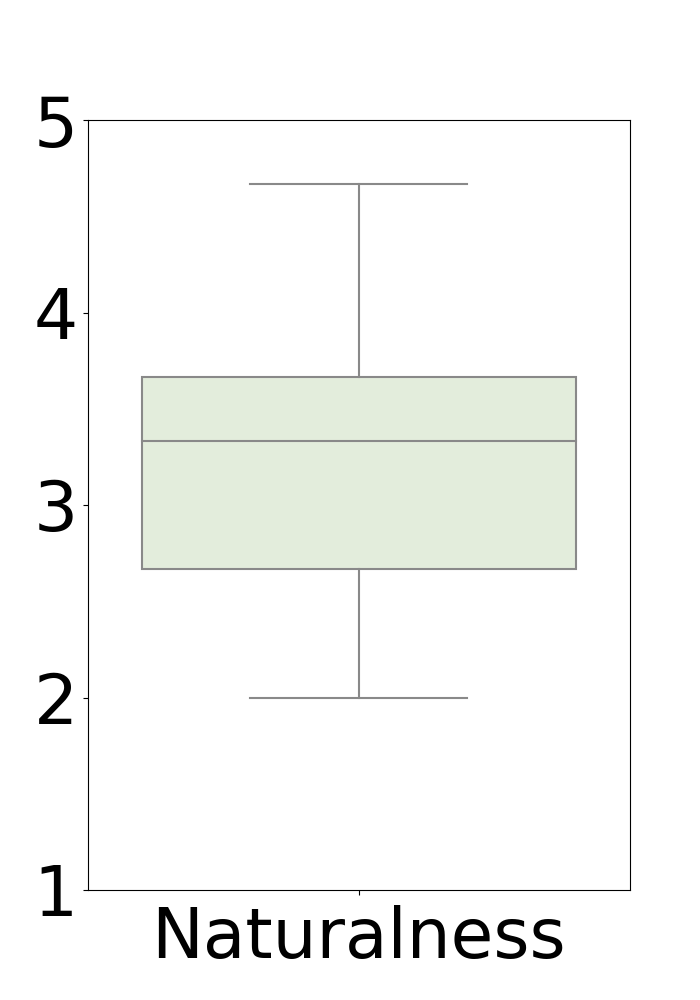}
		\label{subfig:natural}
	\end{minipage}}
\caption{(a) The asterisk shows the Spearman’s correlation between human annotation and our data label, while the bar indicates the QWK scores of MaDSA measured with human labels and synthetic labels respectively. (b) Box plot of the distribution of naturalness scores evaluated by professionals.}
\label{fig:human}
\end{figure}

\subsection{Human Evaluation for Synthetic Data}
To validate our synthetic dataset and confirm its reliability, we conducted the human evaluation with three psychology-domain experts\footnote{We hired professionals with high job success on an online job-search-site Upwork (https://www.upwork.com).} who are highly fluent in English. In detail, we ask them to annotate the PHQ-8 scores for randomly selected 50 dialogue samples within our synthesized dataset. To eliminate individual bias, the value with the most agreement is used as the final expert score.
The asterisk of Figure~\ref{subfig:qwk} shows a strong correlation between the scores from the synthetic data and expert annotation, suggesting the psychological reliability of our synthetic dataset. Moreover, the bar chart in Figure~\ref{subfig:qwk} exhibits similar trends for the QWK values between experts and MaDSA predictions (human-pred) and between the synthetic label and MaDSA (label-pred). 
This suggests the implicit reliability of synthetic data in that expert labels can be used as another testbed.

In addition, we asked psychological experts to evaluate the naturalness (ranging from 1 to 5) of our data samples. Figure~\ref{subfig:natural} demonstrates the effectiveness of our data synthesizing methodology, which reflects turn-wise response similarity between contexts for natural conversation flow.

\subsection{Depression Detection on DAIC-WOZ}
To examine the generality and efficacy of our methodology, we conduct depression detection using the DAIC-WOZ dataset \cite{daicwoz}. It includes labeled a single depression severity based on PHQ-8; however, it is relatively small, comprising 163 training, 56 development, and 56 test samples. Due to its limited size, we utilize a fine-tuned model with our synthetic data, as outlined in Section~\ref{sec:data}. 
Given the absence of PHQ-8 annotations in the test set, our identification results are provided for the development dataset, adhering to the guidelines specified in PHQ-8 \cite{phq8}, whereby cases with a total score of ten or higher are classified as indicative of depression. 
We compare our model with previous study \cite{textbasedet, textbasedet2} that describes the accuracy of the development dataset and use of text-based systems. Table~\ref{tab:daic} shows the performance of our system in detecting depression. While our model is primarily designed to assess multi-faceted depression severity, it is remarkable that our model's capacity for depression classification surpasses that of prior research efforts. This observation implies that our system can potentially contribute to the depression detection task.

\begin{table}[t]
\centering
\caption{Comparison of depression detection accuracy on the DAIC-WOZ development dataset.}
\scalebox{0.8}{
\begin{tabular}{lcc}
\hline
\multicolumn{1}{l|}{Model} &  \multicolumn{1}{c}{Acc. (\%)}  \\ \hline
\multicolumn{1}{l|}{BGRU-ELMo  \cite{textbasedet}
\tablefootnote{The bidirectional GRU based with pre-trained ELMo embedding.}} &  \multicolumn{1}{c}{72.00}    \\ 
\multicolumn{1}{l|}{FS-MT-TXT  \cite{textbasedet2}\tablefootnote{Fully shared multi-task learning with text-based data.} }   & \multicolumn{1}{c}{66.66}  \\ \hline
\multicolumn{1}{l|}{MaDSA (Ours)}           &\multicolumn{1}{c}{74.48}      \\ \hline
\end{tabular}}
\label{tab:daic}
\end{table}

\section{Conclusion}
We introduce a novel task for multi-aspect depression severity assessment via a dialogue system. In particular, we develop a framework that induces highly contextualized responses by utilizing an s emotion classification task to enhance the accuracy of depression severity prediction. To achieve multi-aspect severity assessment, we synthesized the dataset, confirming reliability with human expert evaluations. 
By demonstrating improved predictions in all aspects over the baseline system, our method shows robustness as a foundational system for evaluating depression across multiple dimensions.
Additional experimental results on the DAIC-WOZ suggest an extension of our framework for the depression detection task. 
We hope that our initial efforts could stimulate future interest in the MaDSA task beyond simple depression detection, facilitating detailed explanations for diagnosis.

\bibliographystyle{IEEEtran}
\bibliography{final}

\end{document}